\documentclass[conference]{IEEEtran}
\IEEEoverridecommandlockouts
\usepackage{cite}
\usepackage{amsmath,amssymb,amsfonts}
\usepackage{algorithmic}
\usepackage{graphicx}
\usepackage{textcomp}
\usepackage{xcolor}
\def\BibTeX{{\rm B\kern-.05em{\sc i\kern-.025em b}\kern-.08em
    T\kern-.1667em\lower.7ex\hbox{E}\kern-.125emX}}
\begin{document}

\title{EndoGaussians: Single View Dynamic Gaussian Splatting for Deformable Endoscopic Tissues Reconstruction
}

\author{\IEEEauthorblockN{Yangsen Chen}
\IEEEauthorblockA{\textit{HKUST (Guangzhou)} \\
}
\and
\IEEEauthorblockN{Hao Wang}
\IEEEauthorblockA{\textit{HKUST (Guangzhou)} \\
}
\and
}

\maketitle

\begin{abstract}



The accurate 3D reconstruction of deformable soft body tissues from endoscopic videos is a pivotal challenge in medical applications such as VR surgery and medical image analysis. Existing methods often struggle with accuracy and the ambiguity of hallucinated tissue parts, limiting their practical utility. In this work, we introduce EndoGaussians, a novel approach that employs Gaussian Splatting for dynamic endoscopic 3D reconstruction. This method marks the first use of Gaussian Splatting in this context, overcoming the limitations of previous NeRF-based techniques. Our method sets new state-of-the-art standards, as demonstrated by quantitative assessments on various endoscope datasets. These advancements make our method a promising tool for medical professionals, offering more reliable and efficient 3D reconstructions for practical applications in the medical field.

\end{abstract}


\section{Introduction}

3D Reconstruction of deformable soft body tissues is of great importance for various medical application, such as VR surgery and medical image analysis. In order to precisely reconstruct the deformable soft body tissues from endoscopic videos with single view captures, several previous works have tried to accomplish this task, however, the lack of accuracy and the ambiguity of hallucination part of body tissue are still a big problem, which makes the reconstruction result less trustworthy for practical usages. To further improve the accuracy of the 3D reconstruction of soft body tissue under the static single view RGBD setting, and to improve the reliability and trustworthiness of the 3D reconstruction, we propose EndoGaussians which utilize Gaussian Splatting\cite{gs} as the method for reconstruction. Our method achieves state-of-the-art regarding of several quantitative assessments, such as PSNR, SSIM, LPIPS, etc. and also achieve faster reconstruction speed.

In summary, our contributions are:
\begin{itemize}
\item We propose the first Gaussian Splatting based method for dynamic endoscopic tissues reconstruction.
\item We improve the trustworthiness with the ability of separating hallucination parts on the 3D reconstruction result.
\item We achieve SOTA results on the quantitative assessments on several endoscope datasets.
\end{itemize}

\section{Related Works}

\subsection{Depth estimation based methods}

Previous works \cite{hapnet, huoling} explored the effectiveness of surgical scene reconstruction via depth estimation. Since most of the endoscopes are equipped with stereo cameras, depth can be estimated from binocular vision. 

\subsection{SLAM based methods}

Follow-up SLAM-based methods [23,31,32] fuse depth maps in 3D space to reconstruct surgical scenes under more complex settings. Nevertheless, these methods either hypothesize scenes as static or surgical tools not present, limiting their practical use in real scenarios. 

\subsection{Sparse warp field based methods}

Recent work SuPer \cite{super} and E-DSSR \cite{edssr} present frameworks consisting of tool masking, stereo depth estimation to perform singleview 3D reconstruction of deformable tissues. All these methods track deformation based on a sparse warp field \cite{warp}, which degenerates when deformations are significantly beyond the scope of non-topological changes.

\subsection{NeRF based methods}
With the development of neural radiance field \cite{nerf}, learning based 3d reconstruction has been much more popular, recent works \cite{endonerf, endosurf, lerplane, forplane, based} utilize NeRF for the reconstruction of endoscopic videos. However, due to the implicit representation nature of neural radiance field, user cannot get to know which part of the reconstruction of the tissue is based on real data, and which part of the video is hallucinated.

\section{Methods}
Our framework consists of two steps, Endoscopic Video Inpainting and Single view Dynamic Gaussian Splatting. In the first step, we use video inpainting model to remove the surgical tools from the given endoscopic videos. In the next step, we designed the depth guided dynamic 3d gaussians pipeline for the reconstruction.

\subsection{Problem Setting}

In our research, we focus on reconstructing the surface shape, denoted as $\mathcal{S}$, and texture, $\mathcal{C}$, from a stereo video of deforming tissues. This process is akin to what is seen in EndoNeRF \cite{endonerf}, and involves processing a series of frame data $\left\{\left(\mathbf{I}_i, \mathbf{D}_i, \mathbf{M}_i, \mathbf{P}_i, t_i\right)\right\}_{i=1}^T$, where $T$ represents the total number of frames. For each frame, $\mathbf{I}_i \in \mathbb{R}^{H \times W \times 3}$ is the left RGB image, and $\mathbf{D}_i \in \mathbb{R}^{H \times W}$ is the depth map, both with dimensions height $H$ and width $W$. The foreground mask $\mathbf{M}_i \in \mathbb{R}^{H \times W}$ is used to filter out unnecessary pixels like those from surgical tools, blood, etc.. The projection matrix $\mathbf{P}_i \in \mathbb{R}^{4 \times 4}$ helps in converting 3D coordinates to 2D pixels. While stereo matching, tracking of surgical tools, and pose estimation are significant aspects in clinical applications, our study primarily focuses on 3D reconstruction, assuming that depth maps, foreground masks, and projection matrices are already available through certain software or hardware solutions.

\subsection{Preliminary: Gaussian Splatting}

3D Gaussian Splatting (3D-GS)\cite{gs} is a method used to represent 3D environments with a collection of 3D Gaussians. These Gaussians are defined by a position vector (denoted as \(\boldsymbol{\mu}\) in \(\mathbb{R}^3\)) and a covariance matrix (\(\Sigma\) in \(\mathbb{R}^{3 \times 3}\)). The impact of each Gaussian on a specific point (\(\boldsymbol{x}\)) in 3D space is calculated using the 3D Gaussian distribution equation:

\begin{equation}
   G(\boldsymbol{x})=\frac{1}{(2 \pi)^{3 / 2}|\Sigma|^{1 / 2}} e^{-\frac{1}{2}(\boldsymbol{x}-\boldsymbol{\mu})^T \Sigma^{-1}(\boldsymbol{x}-\boldsymbol{\mu})}
\end{equation}

To ensure that \(\Sigma\) remains positive semi-definite and retains physical significance, it is decomposed into two modifiable components: \(\Sigma=R S S^T R^T\), where \(R\) is a quaternion matrix representing rotation, and \(S\) is a scaling matrix.

Each Gaussian in the system also includes an opacity logit (\(o\)) and several appearance features. These features are represented by \(n\) spherical harmonic (SH) coefficients \(\{c_i \in \mathbb{R}^3 | i=1,2, \ldots, n\}\), where \(n\) equals \(D^2\) and denotes the total count of SH coefficients for degree \(D\). To render a 2D image using 3D-GS, all contributing Gaussians are sequenced for each pixel. The colors from these overlapping Gaussians are combined using the formula:

\begin{equation}
    c=\sum_{i=1}^n c_i \alpha_i \prod_{j=1}^{i-1}\left(1-\alpha_j\right)
\end{equation}

In this formula, \(c_i\) is the color derived from the SH coefficients of the \(i^{\text{th}}\) Gaussian. The value \(\alpha_i\) is obtained by evaluating a 2D Gaussian, with its covariance matrix \(\Sigma^{\prime}\) scaled by opacity. The 2D covariance matrix \(\Sigma^{\prime}\) comes from projecting the 3D covariance \(\Sigma\) onto camera coordinates: \(\Sigma^{\prime}= J W \Sigma W^T J^T\), where \(J\) represents the Jacobian of the affine approximation of the projective transformation and \(W\) is the view transformation matrix.

In the context of point cloud completion, 3D-GS employs a heuristic Gaussian densification strategy to increase Gaussian density. This method relies on the average magnitude of view-space position gradients exceeding a specific threshold. While effective with dense Structure from Motion (SfM) points, this strategy is less successful in fully covering a scene when starting with a very sparse point cloud from sparse-view input images. Additionally, some Gaussians might expand too much, leading to overfitting to training views and poor adaptability to new viewpoints.


\subsection{Endoscopic Video Inpainting}

In our pipeline, the first step is video inpainting. This step aims to remove the surgical tools from the video, because we only need to reconstruct body tissues, we separate this procedure apart from our 3D reconstruction. Previous works \cite{endonerf, endosurf} combines the inpainting of surgical tools into the learning of the neural radiance field, this will introduce hallucination to the model and make training slower, in contrast we separate the inpainting and 3d reconstruction apart, which contains several advantages: firstly, we utilize the flow information of the video, second, we can use explicit representation in the following 3D reconstruction which can explicitly point out the places that was hallucinated during reconstruction, we build up our Single View RGBD Dynamic Gaussians upon the Dynamic Gaussians \cite{dg}. The original Dynamic Gaussians cannot carry out single view reconstruction, therefore we propose several novel components in our gaussian splatting reconstruction pipelines, to make it possible to construct highly accurate soft tissues. The method we used for the video inpainting is Flow-Guided Transformer \cite{fgt}, which is a universal video inpainting model. This model makes use of optical flows to instruct the attention retrieval in transformer for high fidelity video inpainting.

\subsection{Dense Point Cloud Initialization}

The input of our model is the densified point cloud. We argue that under this problem setting, there is no need to use COLMAP for sparse point cloud initialization. Given the RGBD image, we can generate the point cloud using this equation. 

\begin{equation}
\begin{aligned}
X & =\frac{\left(x-c_x\right) \cdot D(x, y) \cdot M(x, y)}{f_x} \\
Y & =\frac{\left(y-c_y\right) \cdot D(x, y) \cdot M(x, y)}{f_y} \\
Z & =D(x, y) \cdot M(x, y)
\end{aligned}
\end{equation}

In this equations, $X, Y$, and $Z$ represent the 3D coordinates of the point in the camera's coordinate system, and $D(x, y)$ is the depth value obtained from the corresponding pixel in the depth image. The mask $M(x, y)$ ensures that invalid pixels do not contribute to the 3D point calculation.

\subsection{Depth Regularization of Dynamic Gaussian Splatting}
\subsubsection{Differentiable Rasterization of Depth}
Drawing inspiration from the work presented in \cite{fsgs}, our approach utilizes Differentiable Depth Rasterization as a means to facilitate backpropagation from the depth prior, enhancing the training process of the Gaussian model. This method is instrumental in capturing the discrepancy between the rendered and estimated depth values through the error signal. In the process of depth rasterization, we employ alpha-blending rendering, a key feature of 3D-GS. This involves aggregating the z-buffer values from the Gaussians that contribute to a pixel, resulting in the depth value calculation as follows:

\begin{equation}
d=\sum_{i=1}^n d_i \alpha_i \prod_{j=1}^{i-1}\left(1-\alpha_j\right)
\end{equation}
   
In this equation, $d_i$ signifies the z-buffer for the $i^{\text{th}}$ Gaussian. The depth loss is enabled by the fully differentiable nature of this implementation, which significantly enhances the match between the rendered and estimated depth values.

\subsubsection{Depth loss}

We use current a learning based stereo matching\cite{unimatch} method to estimation the disparity map of the first frame, then perform parameter search of baseline to fit to the dataset given ground truth depth, with this approach, we can fill the invalid part of the ground truth depth, having better initialization and better depth regularization. We use L1 loss for the depth guidance, but with less weight for the estimated part, while estimating more on the ground truth depth part. We also use an optional loss on the depth smoothness using Huber Loss, which we gain idea from \cite{endosurf}.







\subsubsection{Overall loss function }

For the first frame, we utilize only the image L1 loss and the depth L1 loss:

\begin{equation}
\mathcal{L}=\lambda_1 \underbrace{\|\boldsymbol{C}-\hat{\boldsymbol{C}}\|_1}_{\text {Image L1 Loss }}+\lambda_2 \underbrace{\|\boldsymbol{D}-\hat{\boldsymbol{D}}\|_1}_{\text {Depth L1 Loss }}
\end{equation}

For the following frame, we also assign the same weight for image l1 loss and depth l1 loss as the first frame. While we also adding additional loss to preserve physical correctness following \cite{dg}:

\begin{equation}
\mathcal{L}_{i, j}^{\text {rigid }}=w_{i, j}\left\|\left(\mu_{j, t-1}-\mu_{i, t-1}\right)-R_{i, t-1} R_{i, t}^{-1}\left(\mu_{j, t}-\mu_{i, t}\right)\right\|_2
\end{equation}

\begin{equation}
\mathcal{L}^{\mathrm{rot}}=\frac{1}{k|\mathcal{S}|} \sum_{i \in \mathcal{S}} \sum_{j \in \mathrm{knn}_{i ; k}} w_{i, j}\left\|\hat{q}_{j, t} \hat{q}_{j, t-1}^{-1}-\hat{q}_{i, t} \hat{q}_{i, t-1}^{-1}\right\|_2
\end{equation}

\begin{equation}
\mathcal{L}^{\text {iso }}=\frac{1}{k|\mathcal{S}|} \sum_{i \in \mathcal{S}} \sum_{j \in \mathrm{knn}_{i ; k}} w_{i, j}\left|\left\|\mu_{j, 0}-\mu_{i, 0}\right\|_2-\left\|\mu_{j, t}-\mu_{i, t}\right\|_2\right|
\end{equation}

Therefore, the overall loss for the following frame is:

\begin{equation}
    \begin{aligned}
\mathcal{L} & =\lambda_1 \underbrace{\|\boldsymbol{C}-\hat{\boldsymbol{C}}\|_1}_{\text{Image L1 Loss}} + \lambda_2 \underbrace{\|\boldsymbol{D}-\hat{\boldsymbol{D}}\|_1}_{\text{Depth L1 Loss}} \\
& + \lambda_3 \underbrace{\sum_{i, j} \mathcal{L}_{i, j}^{\text{rigid}}}_{\text{Rigid Loss}} + \lambda_4 \underbrace{\mathcal{L}^{\mathrm{rot}}}_{\text{Rotational Loss}} \\
& + \lambda_5 \underbrace{\mathcal{L}^{\text{iso}}}_{\text{Isometric Loss}}
\end{aligned}
\end{equation}

\subsection{Hallucination Identification}

Under single view RGBD settings, the body tissue occluded by the surgical tools is not always visible, although most of the occluded parts may be visible during some of the frames, it is still not trustworthy to hallucinate them during reconstruction, especially under the scene of medical applications, where accuracy and precision is top one importance.

During the training, the Gaussian Splatting Renderer generates both the RGB image, Depth map, and also the hallucination map. The hallucination map is utilized to compute the hallucination loss, which we defined as the L1 loss between the generated hallucination map and the ground truth tool mask:


\section{Experiments}
Experiments are carried out on two datasets, namely EndoNeRF dataset introuduced in \cite{endonerf} and the SCARED\cite{scared} dataset.

\subsection{Datasets}
Two datasets are used for our experiments.

\subsubsection{EndoNeRF Dataset}
EndoNeRF\cite{endonerf} offers two instances of in-vivo prostatectomy data, which include depth maps that have been estimated using E-DSSR\cite{edssr}, along with tool masks that have been labeled manually.

\subsubsection{SCARED Dataset}
 SCARED \cite{scared} collects the ground truth RGBD images of five porcine cadaver abdominal anatomies.

\subsection{Results}



\begin{tabular}{lcccc}
\hline Methods & \multicolumn{4}{c}{ PSNR $\uparrow$ SSIM $\uparrow$ LPIPS $\downarrow$ FPS $\uparrow$} \\
\hline EndoNeRF\cite{endonerf} & 35.624 & 0.942 & 0.064 & $<0.2$ \\
ForPlane\cite{forplane} & 36.457 & 0.946 & 0.058 & $\sim 1.7$ \\
EndoGS \cite{endogs} & $\mathbf{3 7 . 6 5 4}$ & $\mathbf{0 . 9 6 5}$ & $\mathbf{0 . 0 3 6}$ & $\sim 40$ \\

\hline Ours  & $\mathbf{41. 351}$ & $\mathbf{0 . 9 71}$ & $\mathbf{0 . 0 3 1}$ & $\sim 40$ \\
\hline
\end{tabular}


\begin{figure}[htbp]
    \centering
    \begin{minipage}{0.45\linewidth}
        \includegraphics[width=\linewidth]{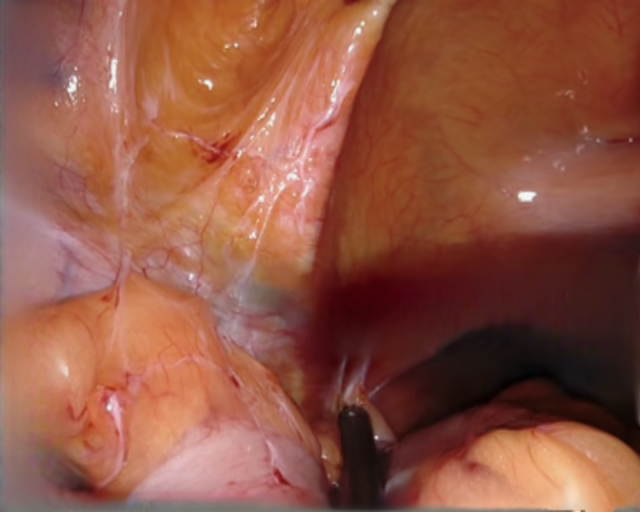}
    \end{minipage}
    \hfill 
    \begin{minipage}{0.45\linewidth}
        \includegraphics[width=\linewidth]{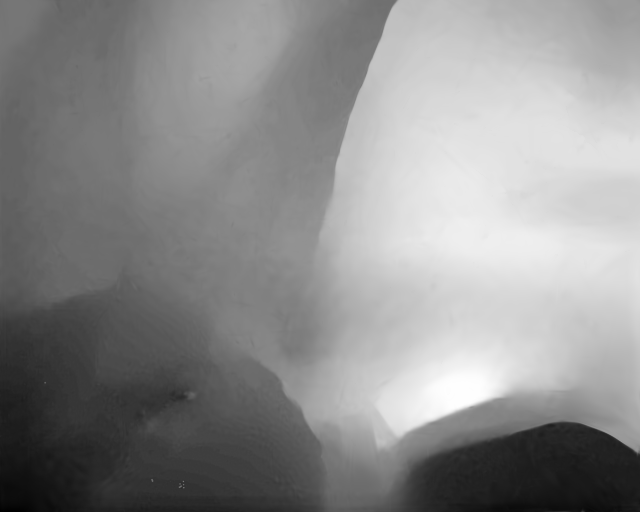}
    \end{minipage}
    \caption{Reconstructed image and depth result of our methods.}
    \label{fig:combined}
\end{figure}



\bibliographystyle{abbrv}
\bibliography{ref}

\end{document}